\documentclass[lettersize,journal]{IEEEtran}

\usepackage{amsmath,amsfonts}
\usepackage{algorithmic}
\usepackage{algorithm}
\usepackage{array}
\usepackage[caption=false,font=normalsize,labelfont=sf,textfont=sf]{subfig}
\usepackage{textcomp}
\usepackage{stfloats}
\usepackage{url}
\usepackage{verbatim}
\usepackage{graphicx}
\usepackage{cite}
\usepackage{booktabs}
\usepackage{xcolor}
\begin{document}

\title{An Empirical Evaluation of Four Off-the-Shelf\\
Proprietary Visual-Inertial Odometry Systems}

\author{Jungha Kim, Minkyeong Song, Yeoeun Lee, Moonkyeong Jung, and Pyojin Kim$^{*}$%
\thanks{All authors are with Department of Mechanical Systems Engineering, Sookmyung Women’s University, Seoul, South Korea. {\tt\small \{alice3071,smk615,snwfry,jmk7791,pjinkim\}@\allowbreak sookmyung.ac.kr}
($^{*}$ Corresponding author: Pyojin Kim)}%
}

\markboth{Journal of \LaTeX\ Class Files,~Vol.~14, No.~8, August~2021}%
{Shell \MakeLowercase{\textit{et al.}}: A Sample Article Using IEEEtran.cls for IEEE Journals}


\maketitle

\begin{abstract}
Commercial visual-inertial odometry (VIO) systems have been gaining attention as cost-effective, off-the-shelf six degrees of freedom (6-DoF) ego-motion tracking methods for estimating accurate and consistent camera pose data, in addition to their ability to operate without external localization from motion capture or global positioning systems.
It is unclear from existing results, however, which commercial VIO platforms are the most stable, consistent, and accurate in terms of state estimation for indoor and outdoor robotic applications.
We assess four popular proprietary VIO systems (Apple ARKit, Google ARCore, Intel RealSense T265, and Stereolabs ZED 2) through a series of both indoor and outdoor experiments where we show their positioning stability, consistency, and accuracy.
We present our complete results as a benchmark comparison for the research community.
\end{abstract}


\section{Introduction}
\label{sect:intro}

\IEEEPARstart{T}{his} article presents a benchmark comparison of off-the-shelf proprietary visual-inertial odometry (VIO) systems used for autonomous navigation of robotic applications, which are the process of determining the position and orientation of a camera-inertial measurement unit (IMU)-rig in 3D space by analyzing the associated camera images and IMU data.
As the VIO research has reached a level of maturity, there exist several open published VIO methods such as MSCKF~\cite{mourikis2007multi}, OKVIS~\cite{leutenegger2013keyframe}, VINS-Mono~\cite{qin2018vins}, and many commercial products utilize closed proprietary VIO algorithms such as Apple ARKit~\cite{apple2022arkit}, Google ARCore~\cite{google2022arcore} that offer off-the-shelf VIO pipelines which can be employed on an end-user's system of choice.

The current research studies provide some comparative experiments on the performance of the popular VIO approaches, however, they consider only a subset of the existing open-source and proprietary VIO algorithms, and conduct insufficient performance evaluation only on publicly-available datasets rather than indoor and outdoor challenging real-world environments.
In particular, although commercial VIO systems (Intel T265, Stereolabs ZED 2) play an important role in the several DARPA challenges~\cite{rouvcek2019darpa, root2021fast} and many commercial products or apps (Pokémon GO, IKEA Place AR), there is a lack of research for benchmarking the positioning accuracy of these closed proprietary VIO platforms.

\begin{figure}[!t]
\centering
\includegraphics[width=\linewidth]{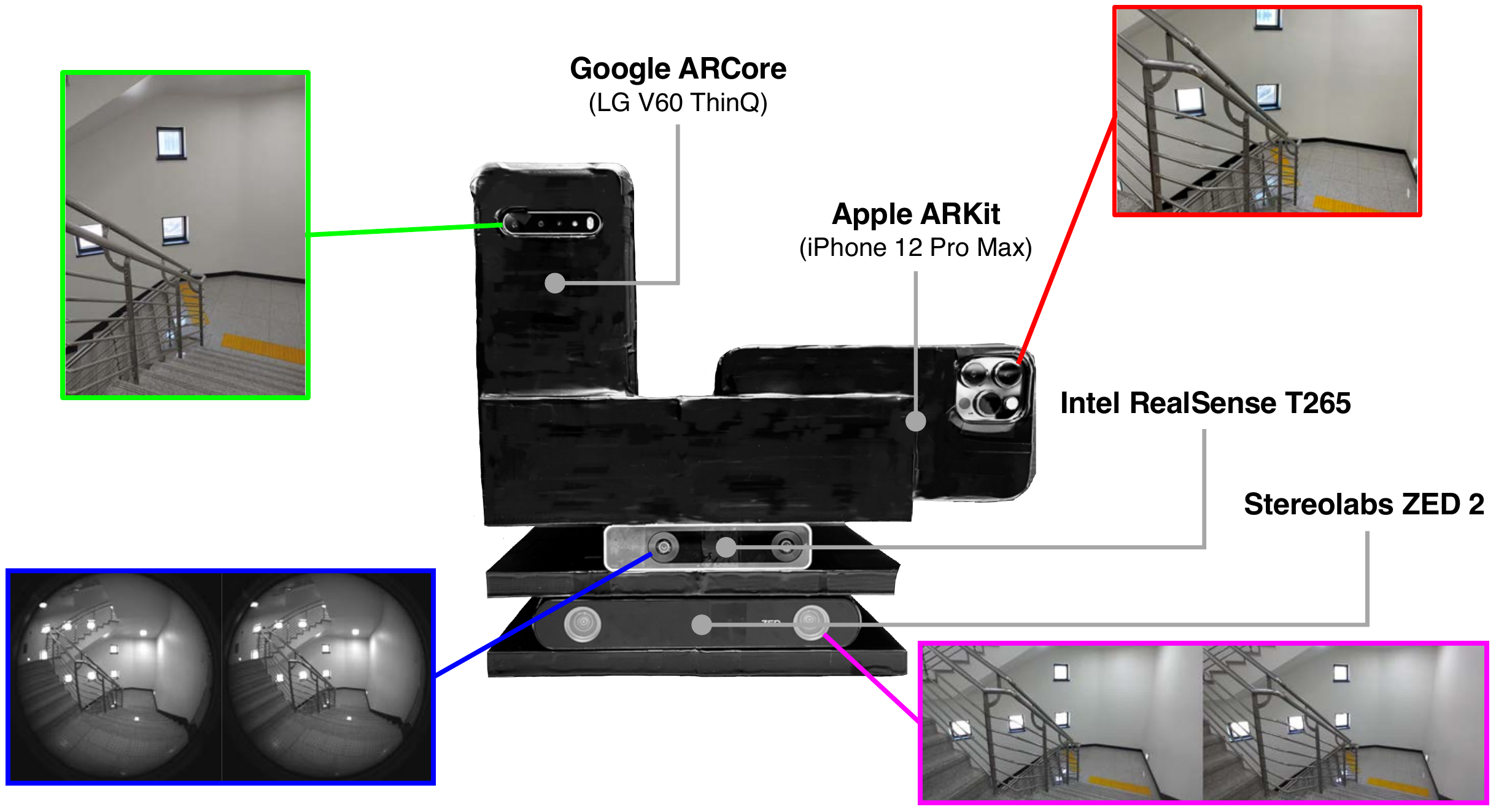}
\vspace{-6mm}
\caption{The custom-built capture rig for benchmarking 6-DoF motion tracking performance of Apple ARKit (iPhone 12 Pro Max), Google ARCore (LG V60 ThinQ), Intel RealSense T265, and Stereolabs ZED 2.}
\vspace{-4mm}
\label{figure_capture_rig}
\end{figure}

\begin{figure*}[!h]
\centering
\includegraphics[width=\linewidth]{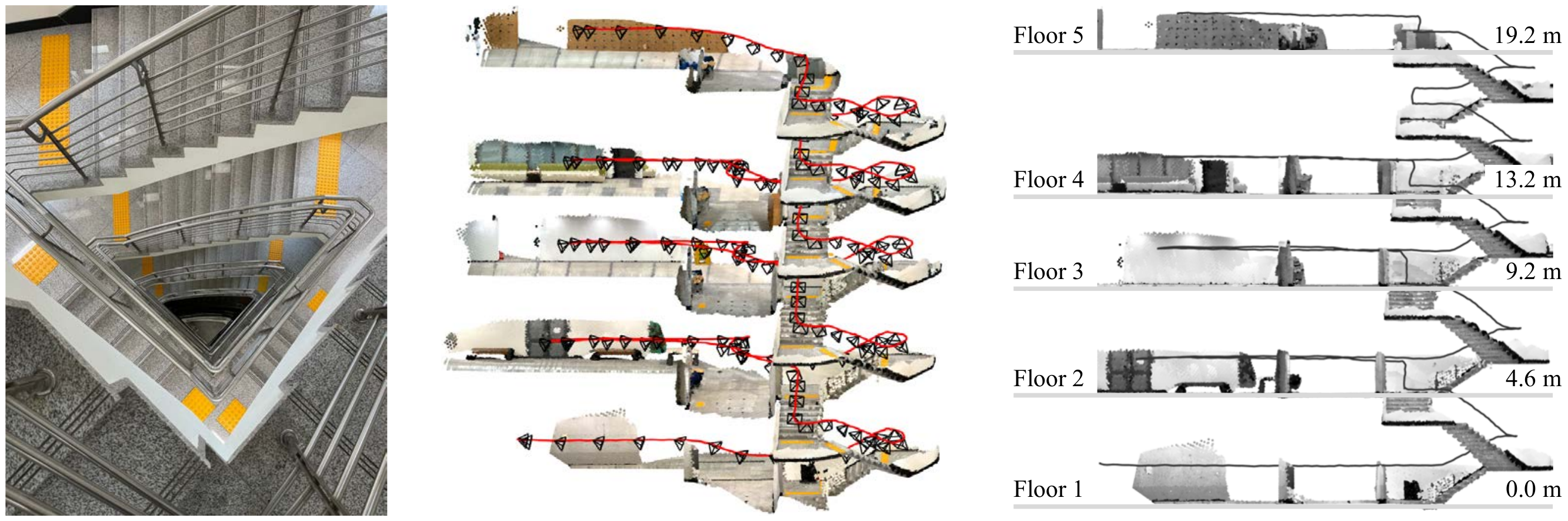}
\vspace{-6mm}
\caption{Accumulated 3D point cloud (middle) with the estimated 6-DoF trajectory (red) from Apple ARKit in multi-floor environments.
We capture the 6-DoF camera poses and 3D points while climbing the multi-story stairs (left).
Among the four proprietary VIO systems, Apple ARKit shows the most consistent and accurate 6-DoF motion tracking results, consistently reconstructing the 3D geometry of stairs and hallways.
The Apple ARKit track (red) and 3D reconstruction results have a similar shape as the ground-truth blueprint of a building (right).}
\vspace{-4mm}
\label{figure_front_page}
\end{figure*}

The motivation of this paper is to address this deficiency by performing a comprehensive evaluation of off-the-shelf commercially-available VIO systems in challenging indoor and outdoor environments as shown in Fig.~\ref{figure_front_page}.
This is the first comparative study on four popular proprietary VIO systems in six challenging real-world environments, both indoors and outdoors.
Especially, we select the following four proprietary VIO systems that are frequently used in autonomous driving robotic applications:
\begin{itemize}
\item[$\bullet$] Apple ARKit~\cite{apple2022arkit} - Apple's augmented reality (AR) platform, which includes filtering-based VIO algorithms~\cite{flint2018visual} to enable iOS devices to sense how they move in 3D space.
\item[$\bullet$] Google ARCore~\cite{google2022arcore} - Google's AR platform utilizing a multi-state constraint Kalman filter (MSCKF) style VIO algorithm~\cite{mourikis2007multi, mourikis2009vision}, called concurrent odometry and mapping (COM)~\cite{nerurkar2020system}.
\item[$\bullet$] Intel RealSense T265~\cite{intel2022t265} - a stand-alone VIO and simultaneous localization and mapping (SLAM) tracking device developed for use in robotics, drones, and more, with all position computations performed on the device.
\item[$\bullet$] Stereolabs ZED 2~\cite{stereolabs2022zed2} - a hand-held stereo camera with built-in IMU for neural depth sensing and visual-inertial stereo, requiring an external NVIDIA GPU to obtain the 6-DoF camera poses.
\end{itemize}

We do not consider open-source published VIO methods and non-inertial visual simultaneous localization and mapping (SLAM) algorithms, for example ROVIO~\cite{bloesch2015robust}, VINS-Mono~\cite{qin2018vins}, ORB-SLAM~\cite{mur2017orb}, and DSO~\cite{engel2017direct}.
We focus on the off-the-shelf commercial VIO/SLAM products that might be of interest to more researchers and engineers because open published VIO algorithms are relatively difficult to understand and operate, and their comparisons are made in the literature~\cite{delmerico2018benchmark, campos2021orb} to some extent.

Our experiments are conducted in six challenging indoor and outdoor environments with the custom-built test rig equipped with the four VIO devices as illustrated in Fig.~\ref{figure_capture_rig}.
Our test sequences contain long and narrow corridors, large open spaces, repetitive stairways, an underground parking lot with insufficient lighting, and about 3.1 kilometers of a vehicular test in complex urban traffic environments.

Our goal is to provide a thorough benchmark of closed proprietary VIO systems, in order to provide a reference for researchers on VIO products, as well as readers who require an off-the-shelf 6-DoF state estimation solution that is suitable for their robotic platforms and autonomous vehicles.




\section{Related Work}
\label{sect:relatedwork}

Despite proprietary VIO systems being utilized in many products and areas for industrial usage (e.g., for building an accurate indoor map, as a precise positioning system, etc.), there is no benchmark study that satisfies our proposed goals.
While comprehensive comparisons of open-source published VIO methods exist~\cite{delmerico2018benchmark}, they focus only on evaluating the popular academic VIO algorithms on the EuRoC micro aerial vehicle dataset~\cite{burri2016euroc}, and do not cover off-the-shelf proprietary VIO systems and various indoor and outdoor environments.
Although ADVIO~\cite{cortes2018advio} presents a VIO comparison including three proprietary platforms and two academic approaches, its main contribution is to develop a set of RGB and IMU smartphone datasets, not a performance evaluation between the proprietary VIO platforms.
In \cite{alapetite2020comparison, ouerghi2020comparative}, some comparative studies of proprietary VIO systems have been performed, but they consider only a few proprietary VIO platforms.
Performance evaluation is only conducted in a simple 2D indoor environment with a short camera moving distance.

Since we focus on the 6-DoF positioning accuracy of the proprietary VIO systems, we can instead consider the existing results relevant to this problem.
The proposed VIO approach in \cite{ling2018modeling} compares to Google ARCore and VINS-Mono~\cite{qin2018vins}, but only on a few indoor sequences with very little camera movement.
The evaluation framework in \cite{gumgumcu2019evaluation} assesses the 6-DoF motion tracking
performance of ARCore with the ground truth under several circumstances, but they lack comparative results for other proprietary VIO systems such as ARKit and T265, and detailed analyses are performed only for ARCore.

Most important is that no existing work considers an indoor/outdoor performance evaluation for four popular proprietary VIO systems that are frequently deployed on robotic applications, AR/VR apps, and industrial usages.
Our test sequences are authentic and illustrate realistic use cases, containing challenging environments with scarce or repetitive visual features, both indoors and outdoors, and varying motions from walking to driving camera movements.
They also include rapid rotations without translation as they are problematic motions for many VIO/SLAM algorithms.


\section{Visual-Inertial Odometry Systems}
\label{sect:viosystems}

We briefly summarize the primary features of four off-the-shelf proprietary VIO systems based on data published on the relevant official websites, papers, GitHub, and patent documents, and how to collect 6-DoF pose estimates from each VIO mobile device.
Since most proprietary VIO/SLAM platforms are all closed-source, we do not cover the detailed VIO academic backgrounds and implementations.

\subsection{Apple ARKit}
\label{subsec:applearkit}
Apple ARKit~\cite{apple2022arkit} is Apple's augmented reality (AR) software framework, which includes a tightly-coupled filtering-based VIO algorithm similar to the MSCKF~\cite{mourikis2007multi} to enable iOS devices to sense how they move in 3D space.
It contains a sliding window filter, bundle adjustment, motion/structure marginalization modules~\cite{flint2018visual}, and is expected to be applied to various robotic applications such as Apple Glasses and Car in the future, not just for iPhone and iPad, which is why we conduct vehicle tests in this benchmark.
We develop a custom iOS data collection app\footnote{\url{https://github.com/PyojinKim/ios_logger}} for capturing ARKit 6-DoF camera poses, RGB image sequences, and IMU measurements using an iPhone 12 Pro Max running iOS 14.7.1.
It saves the pose estimates as a translation vector and a unit quaternion at 60 Hz, and each pose is expressed in a global coordinate frame created by the phone when starting iOS data collection.
Although there are various iPhone and iPad models, the core VIO algorithm in ARKit is the same, thus we empirically confirm that there is little difference in the VIO performance of each device.

\subsection{Google ARCore}
\label{subsec:googlearcore}
ARCore~\cite{google2022arcore} is Google’s platform for building AR experiences utilizing the multi-state constraint Kalman filter (MSCKF) style VIO/SLAM algorithms~\cite{mourikis2009vision, nerurkar2014c} with many subsequent variations, called concurrent odometry and mapping (COM)~\cite{nerurkar2020system}.
ARCore is a successor to Google Project Tango~\cite{marder2016project}, and is currently applied only to Android OS smartphones, but it would be extended to various robotic platforms such as Google Wing, Maps, and Waymo, which is why we evaluate ARCore in a large-scale outdoor sequence of about 3.1 kilometers of a vehicular test.
We build a custom Android OS app based on Google’s ARCore example\footnote{\url{https://github.com/rfbr/IMU_and_pose_Android_Recorder}} to acquire ARCore 6-DoF camera poses and IMU measurements at 30 Hz with an LG V60 ThinQ running Android 10.0.0 and ARCore 1.29.
Although there are various Android OS devices such as Samsung Galaxy and Google Pixel, smartphones on the list\footnote{\url{https://developers.google.com/ar/devices}} certified by Google demonstrate similar motion tracking performance regardless of device model.

\subsection{Intel RealSense T265}
\label{subsec:intelT265}
Intel RealSense T265 is a hassle-free stand-alone VIO/SLAM device to track its own position and orientation in 3D space.
The embedded processor, vision processing unit (VPU), runs the entire VIO algorithm onboard, analyzes the image sequences from stereo fisheye cameras and fuses all sensor information together.
Since the T265 VIO algorithm runs on the device itself without using the resource of the host computer, it is widely used as a 6-DoF positioning sensor in 3D space for various robotic applications such as DARPA challenges~\cite{rouvcek2019darpa} and autonomous flying drones~\cite{bonatti2020learning}.
We collect the 6-DoF motion tracking results at 200 Hz using Intel RealSense SDK 2.0\footnote{\url{https://github.com/IntelRealSense/librealsense}}, and save the T265 6-DoF camera poses by connecting it to an Intel NUC mini PC.

\subsection{Stereolabs ZED 2}
\label{subsec:stereolabszed2}
Stereolabs ZED 2 is a hand-held stereo camera with built-in IMU for neural depth sensing, 6-DoF VIO/SLAM, and real-time 3D mapping.
Stereolabs has not made their VIO/SLAM algorithm public, and the description of the VIO algorithm is relatively vague compared to other proprietary VIO systems.
It is one of the popular stereo camera sensors for various robotic applications such as drone inspection~\cite{fan2019real}, but has the disadvantage of requiring an external NVIDIA GPU to perform positional tracking and neural depth sensing.
We develop a program to collect the ZED 2 6-DoF camera poses at 30 Hz based on ZED SDK 3.5.2\footnote{\url{https://www.stereolabs.com/developers/release/}} on an NVIDIA Jetson Nano onboard computer.

\section{Experiments}
\label{sect:experiments}

\begin{figure}[!t]
\centering
\includegraphics[width=\linewidth]{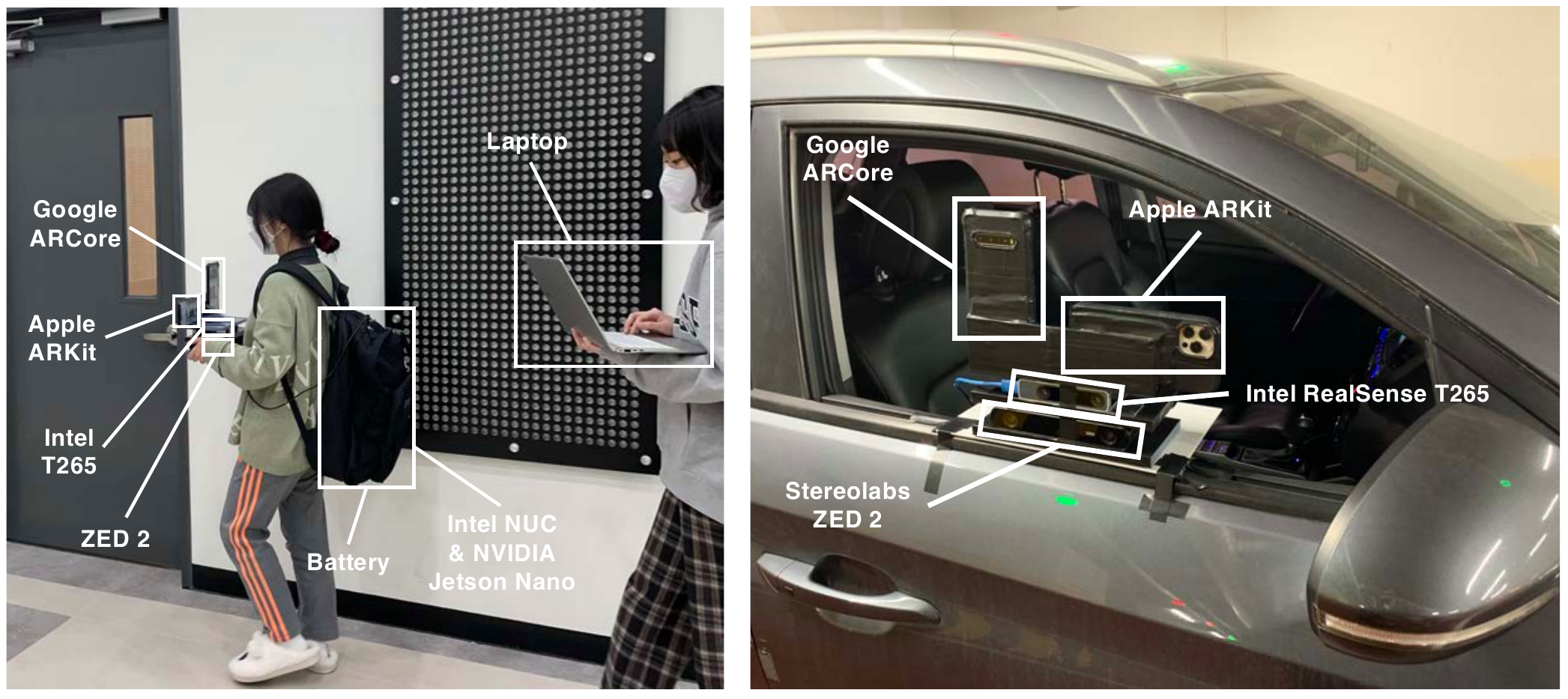}
\vspace{-6mm}
\caption{We carry the capture rig by hand, and store the onboard computers and batteries to collect the motion data indoors (left).
In the outdoor vehicular tests, we fix the capture rig to the front passenger seat (right).}
\vspace{-5mm}
\label{figure_external_view}
\end{figure}

We evaluate four proprietary VIO systems with the four devices (iPhone 12 Pro Max, LG V60 ThinQ, Intel T265, ZED2) attached to the custom-built capture rig as shown in Fig.~\ref{figure_capture_rig} and Fig.~\ref{figure_external_view} on the large-scale challenging indoor and outdoor environments, both qualitatively and quantitatively.
Indoors, we record the motion data by a walking person, and outdoors, the data is collected by rigidly attaching the capture rig to a car in Fig.~\ref{figure_external_view}.
We save the 6-DoF pose estimates of ARKit and ARCore through the custom apps on each smartphone device, and record the moving trajectories of T265 and ZED2 in the Intel NUC and NVIDIA Jetson Nano onboard computers.
We maintain the default parameter settings of each VIO platform, and deactivate all capabilities related to SLAM (e.g., loop closure) for a fair comparison between each VIO system.
Furthermore, in order to interpret the motion tracking results in the same reference coordinate frame, we calibrate the intrinsic and extrinsic parameters of all cameras by capturing multiple views of a checkerboard.

\begin{figure*}[!h]
\centering
\includegraphics[width=\linewidth]{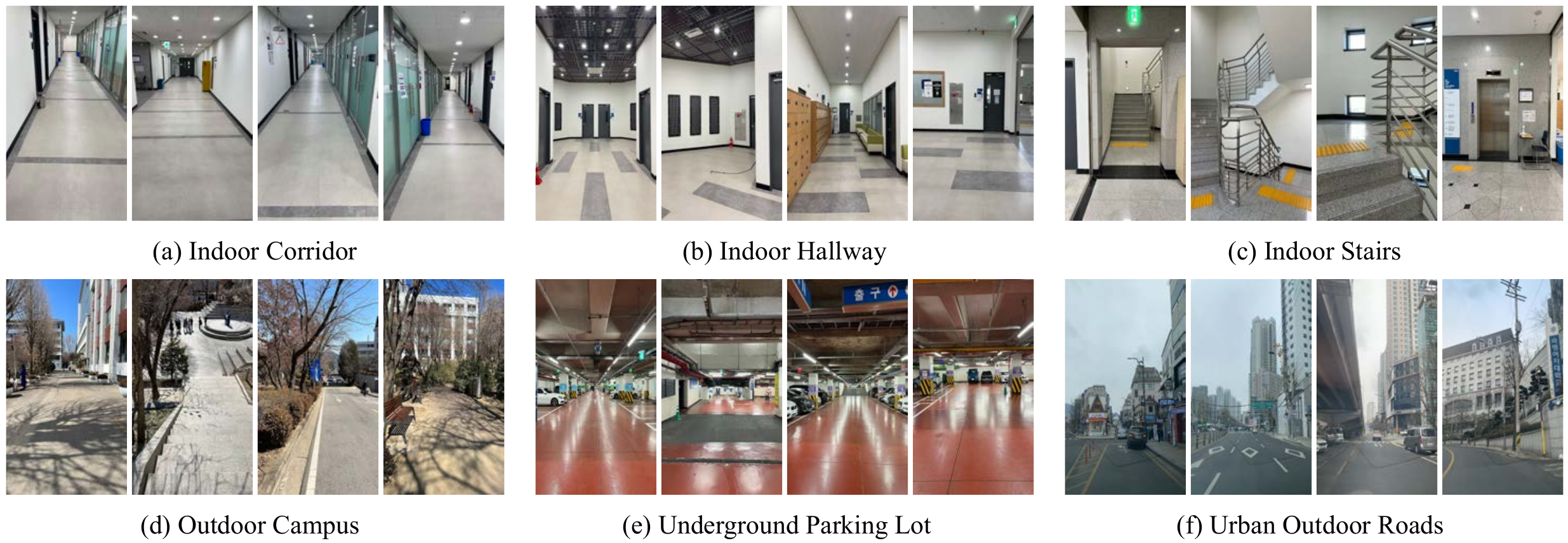}
\caption{Example image frames from indoor and outdoor benchmark datasets.
The top row represents three indoor sequences by foot including long corridors (a), open hallway spaces (b), and repetitive stairs (c) from a university building.
We acquire the camera motion data in the outdoor campus on foot (d), and through a car in the underground parking lot (e) and urban outdoor roads (f).}
\label{figure_example_frames}
\end{figure*}

Our benchmark dataset contains various indoor and outdoor sequences in six different locations, and the total length of each sequence ranges from 83 m to 3051 m, which are primarily designed for benchmarking medium and long-range VIO performance.
There are three indoor and three outdoor sequences, and all indoor sequences are captured in a 7-story building in the university campus including long corridors, open hallway spaces, and stair climbs as shown in the top row of Fig.~\ref{figure_example_frames}.
The indoor cases are as realistic as possible containing repetitive motion in stairs, temporary occlusions, and areas lacking visual features.
The bottom row of Fig.~\ref{figure_example_frames} illustrates example frames from three outdoor sequences acquired in the outdoor university campus, underground parking lot, and urban outdoor roads.
In order to evaluate the performance of each VIO system quantitatively without an external motion capture system, we coincide the start and end points of the movement trajectories in all experiments, and measure the final drift error (FDE) metric, which is the end point position error in meter.
We report the quantitative evaluation results of four VIO systems in Table~\ref{table_evaluation_FDE}.
The smallest end point position error for each sequence is indicated in bold.
The ideal FDE value (the ground-truth path) should be 0, and a large FDE value denotes an inaccurate position estimate since we define the starting point of movement as the origin.
In addition, by overlaying the estimated VIO trajectories on the floorplan of the building or Google Maps, we evaluate the consistency, stability, and reliability of each VIO system qualitatively.

\subsection{Indoor Long Corridors and Open Hallway Sequences}

\begin{figure*}[!h]
\centering
\includegraphics[width=\linewidth]{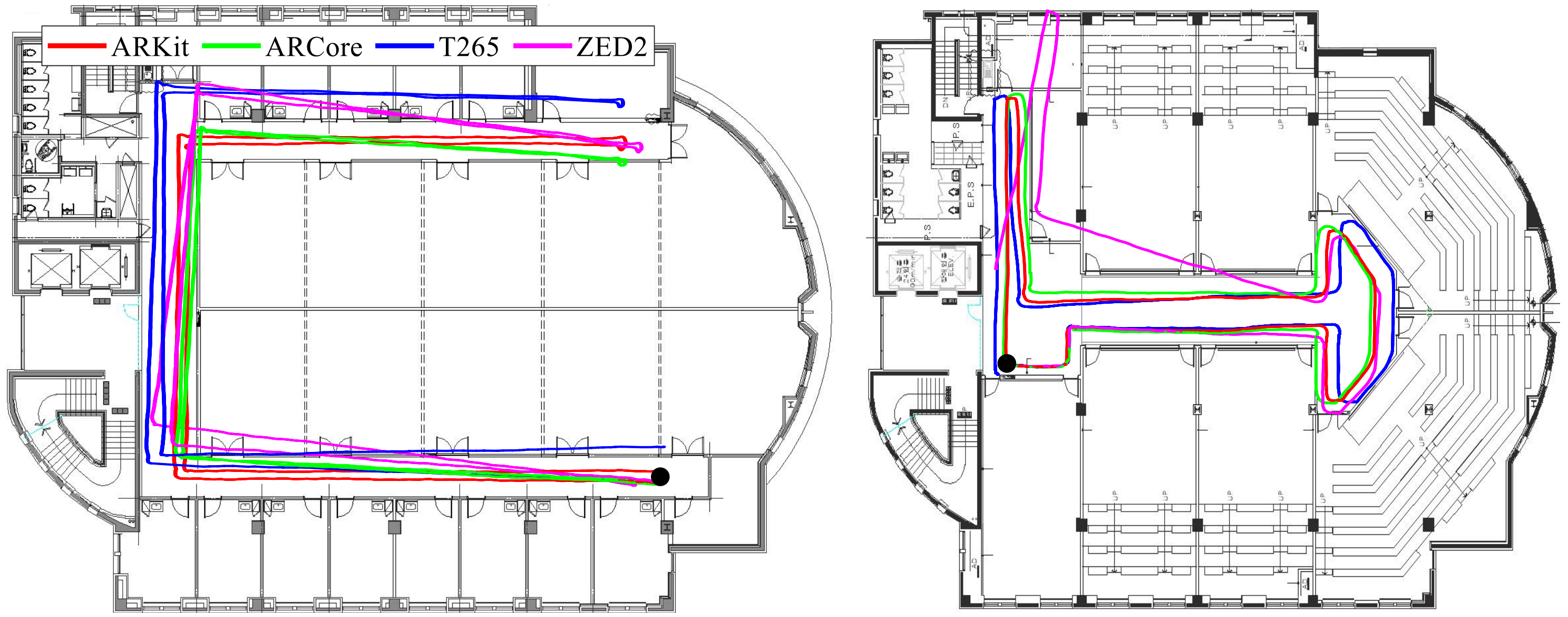}
\vspace{-5mm}
\caption{Estimated trajectories with four proprietary VIO systems in a long U-shape corridor (left) and open hallway space (right) sequences.
We start and end at the same point marked in the black circle to evaluate the loop closing performance of tested commercial VIO systems.
The estimated paths for ARKit (red) match the building floorplan most consistently, and only the starting and ending points of ARKit nearly meet; for the others, they do not.}
\vspace{-3mm}
\label{figure_indoor_corridor_hallway}
\end{figure*}

\begin{table}[!t]
\caption{Evaluation Results (FDE) of Four Proprietary VIO Systems}
\renewcommand{\arraystretch}{1.1}
\label{table_evaluation_FDE}
\centering
\resizebox{8.4cm}{!}{%
\begin{tabular}{l|cccc|c}
  \toprule
Experiment       &     ARKit    &    ARCore   &     T265    &     ZED 2   &   Length (m) \\
  \midrule
Indoor Corridor  &     0.79     &\textbf{0.12}&     1.88    &     1.44    &    145.21    \\
Indoor Hallway   &     0.14     &\textbf{0.09}&     0.61    &     4.58    &     83.98    \\
Indoor Stairs    &\textbf{0.19} &     3.98    &     1.49    &     4.76    &    114.13    \\
  \midrule
Outdoor Campus   &     2.01     &\textbf{0.07}&     4.08    &    206.38   &    513.81    \\
Parking Lot      &\textbf{0.26} &      1.14   &     9.01    &     10.85   &    446.26    \\
Outdoor Roads    &\textbf{2.68} &    140.08   &   $\times$  &    409.25   &   3051.61    \\
  \bottomrule
\end{tabular}%
}
\end{table}

We evaluate four VIO systems in a long U-shape corridor and open hallway spaces easily found in typical office and university buildings as shown in Fig.~\ref{figure_indoor_corridor_hallway}.
Fig.~\ref{figure_example_frames} (a) and (b) illustrate example frames from both locations.
The trajectories of these sequences are approximately 145 and 84 meters, and include 5 and 11 pure rotational movements and difficult textures.
In a long U-shape corridor and open hallway sequences, the start and end points of ARKit (red) meet at the black circle without a severe rotational drift while maintaining the orthogonality and scale of the estimated trajectory well compared with the floorplan.
Although ARCore (green) shows the most accurate results in terms of the FDE metric in Table~\ref{table_evaluation_FDE}, the estimated VIO trajectory does not match the floorplan well.
Intel T265 (blue) estimates accurate 3-DoF rotational motion well, but there is a problem with the scale of the moving trajectory compared to the floorplan, showing a little larger trajectory than the actual movements.
ZED2 (magenta) presents the most inaccurate and inconsistent positioning performance among the four VIO methods as the rotational motion drift error gradually accumulates over time.
Overall, the estimated VIO trajectories by ARKit (red) are the most similar and consistent motion tracking results to the actual movements following the shape of the corridor on the floorplan.

\subsection{Indoor Multistory Stairs Sequence}

\begin{figure}[!t]
\centering
\includegraphics[width=\linewidth]{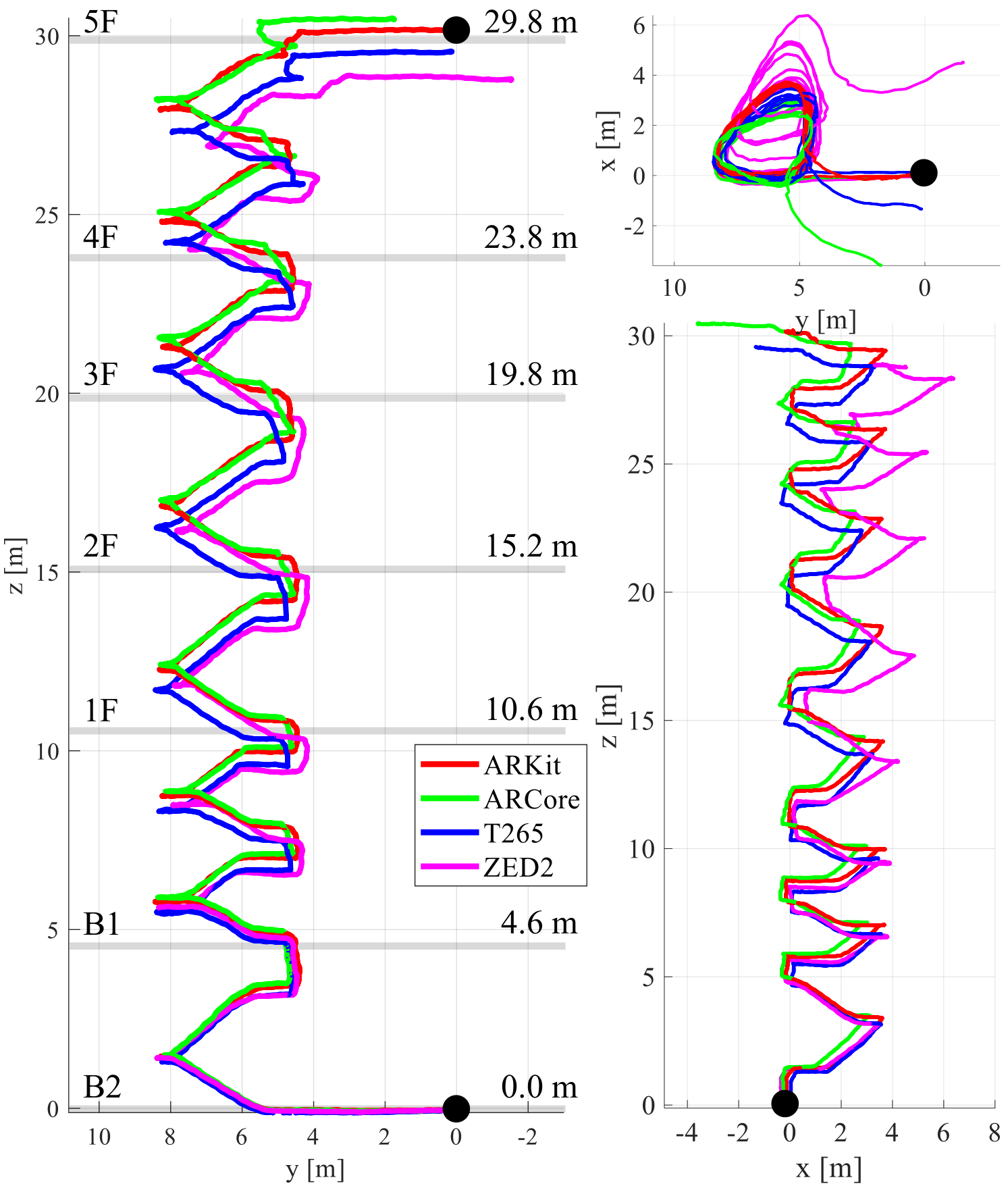}
\vspace{-5mm}
\caption{Comparison of four VIO systems on multi-story stairs from the 2nd basement floor (B2) to the 5th floor (5F).
It shows the side (left), front (right-bottom), and top (right-top) views of the estimated VIO trajectories.
ARKit has the most consistent camera motions along with the shape of the stairs, and only matches the start and end points marked in the black circle.}
\label{figure_indoor_stairs}
\end{figure}

We perform a comparative experiment in a multi-floor staircase environment with the 114 m trajectory going up the stairs from the 2nd basement floor (B2) to the 5th floor (5F) of a building in Fig.~\ref{figure_indoor_stairs}.
The repetitive rotational motion included in the 3D trajectory of climbing the stairs makes VIO positioning challenging.
Fig.~\ref{figure_example_frames} (c) shows example frames from multistory stair sequence.
In the top view (xy-plane), we start and end at the same points marked in the black circle to check loop closing in the estimated VIO trajectories.
ARKit (red) has the best performance; the top and side views of ARKit (red) show the overlapped, consistent 6-DoF motion tracking results while other VIO systems gradually diverge from the initially estimated loop.
With ARKit (red), the starting and ending points in xy-plane (top view) nearly match; for the others, they do not.
The final drift error (FDE) of ARKit in xy-plane is 0.19 m, while ARCore, T265, and ZED2 are 3.98 m, 1.49 m, and 4.76 m, respectively.
In particular, ZED2 (magenta) has the most severe trajectory distortion in the z-axis direction (height) among the four VIO systems.
Fig.~\ref{figure_indoor_stairs} illustrates the side and front views of the stairway with the paths from four VIO devices, showing high consistency of ARKit (red) compared to other VIO platforms.
It is noteworthy that the height of each floor estimated by ARKit and the actual height (the ground-truth) from the building blueprint are approximately identical.

\subsection{Outdoor University Campus Sequence}

\begin{figure*}[!h]
\centering
\includegraphics[width=\linewidth]{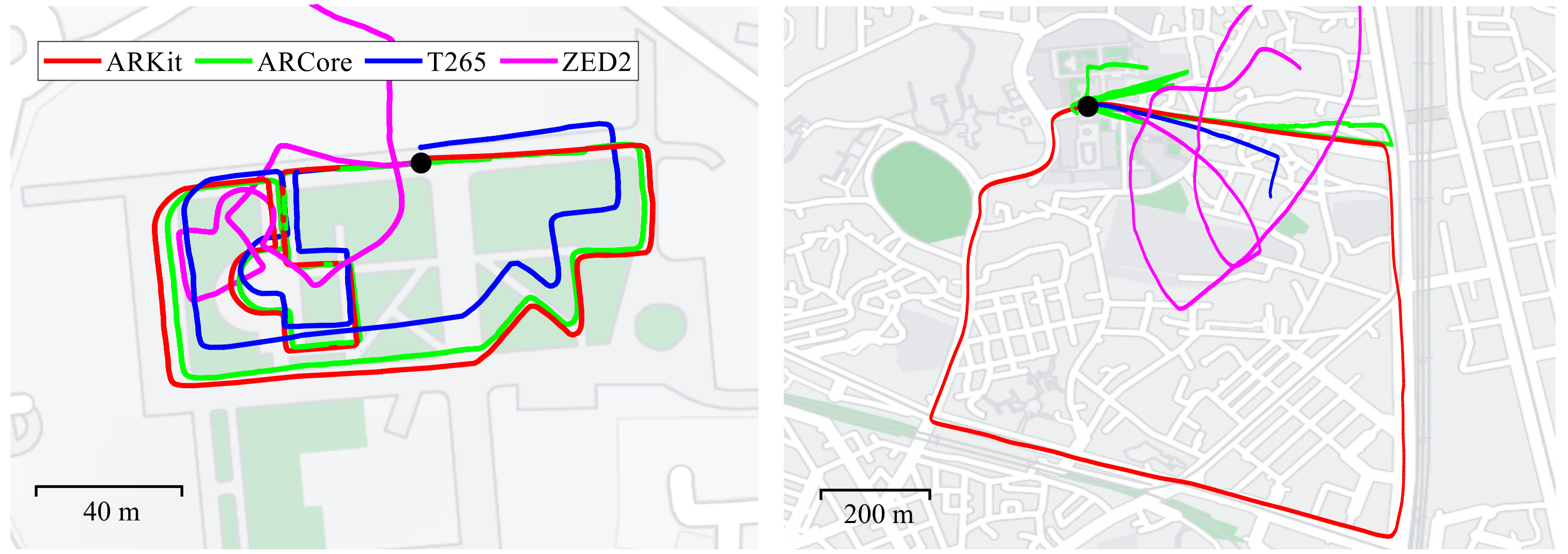}
\caption{Estimated motion trajectories of four proprietary VIO systems in an outdoor campus (left) and urban outdoor roads (right) sequences overlaid on Google Maps.
We start and end at the same point marked in the black circle to check loop closing performance.
ARKit (red) tracks the 6-DoF camera poses well following the shape of the roads on Google Maps most consistently and accurately.
Only ARKit (red) is able to produce stable motion tracking performance even when driving a vehicle over 60 km/h (right).}
\vspace{-4mm}
\label{figure_outdoor_campus_roads}
\end{figure*}

We choose an outdoor location in the university campus approximately 513 m to determine which VIO system works well in the environment of the rapid change of the topography, and the narrow road returning as shown in the left of Fig.~\ref{figure_outdoor_campus_roads}.
Example frames are shown in Fig.~\ref{figure_example_frames} (d).
It shows the resulting 6-DoF trajectories from four VIO platforms overlaid on Google Maps, demonstrating that the start and end points of ARKit (red) and ARCore (green) meet while matching well with the shape of the roads shown on Google Maps.
The shape of the estimated trajectory of T265 (blue) is very similar to ARKit and ARcore's results, however, the scale of the estimated path of T265 is smaller than the actual movements.
T265 suffers from a scale inconsistency problem, which is generally observed in monocular visual odometry configuration.
The orthogonality of ZED2 (magenta) is broken due to its inaccurate rotation estimation, showing the most severe distortion of the actual movement trajectory among the four VIO systems as shown in the left of Fig.~\ref{figure_outdoor_campus_roads}.

\subsection{Outdoor Urban Roads and Parking Lot Sequences}

\begin{figure}[!t]
\centering
\includegraphics[width=\linewidth]{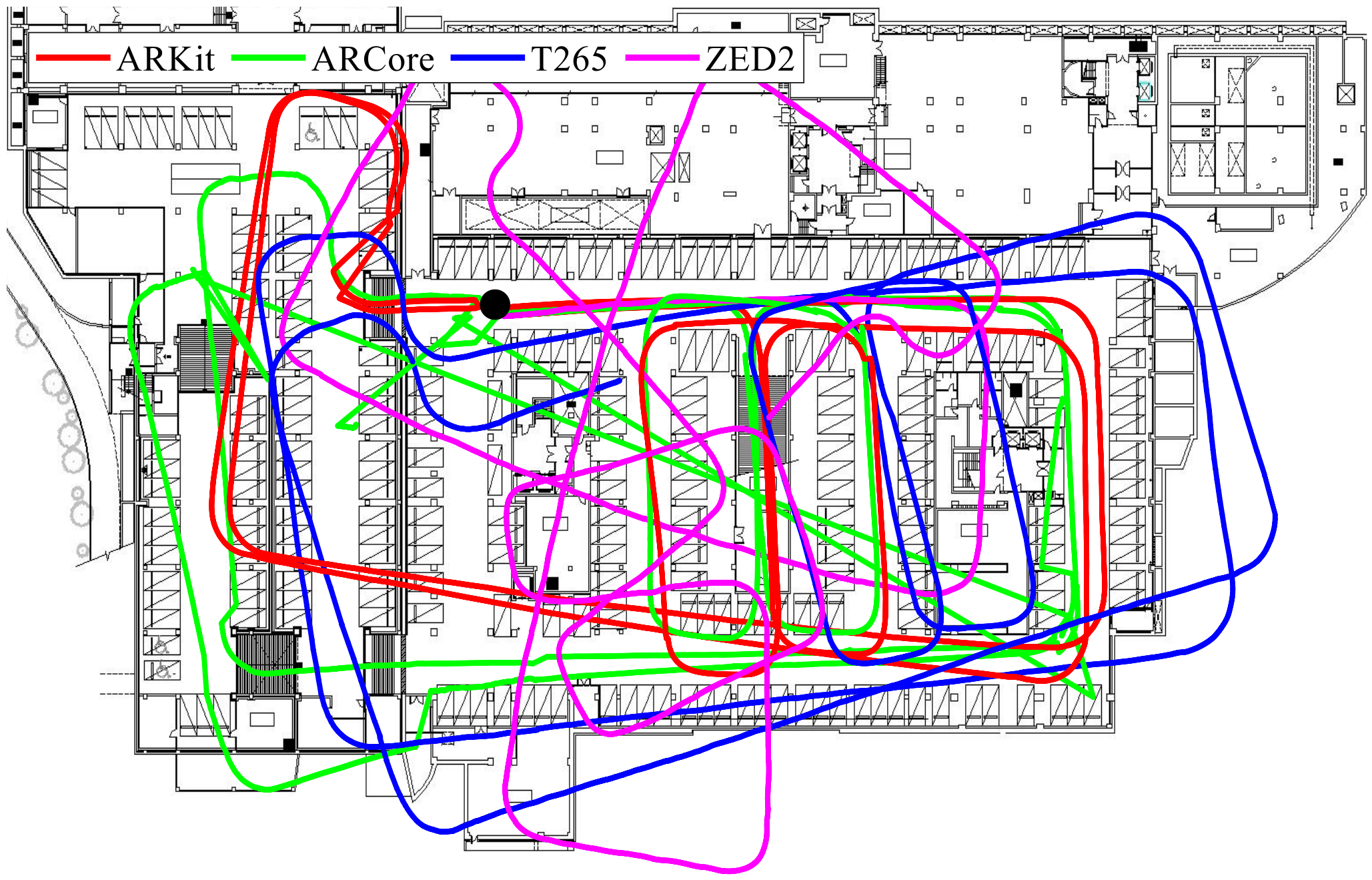}
\caption{Example paths in the underground parking lot overlaid on the floorplan to evaluate the consistency and accuracy.
The trajectories of ARKit (red) overlap significantly, but the paths of the other VIO devices suffer from a rotational drift, showing inaccurate and inconsistent positioning results.}
\label{figure_underground_parking_lot}
\end{figure}

We perform an outdoor vehicle driving experiment with a mileage of approximately 3 km by attaching the capture rig to the vehicle as shown in the right of Fig.~\ref{figure_outdoor_campus_roads}.
Fig.~\ref{figure_example_frames} (e) and (f) show example frames from the underground parking lot and urban outdoor roads.
We acquire the motion data while driving on public automobile roads near Seoul Station in Seoul, and there are plenty of moving people, cars, and occasional large vehicles visible in the outdoor environments, which makes motion tracking of VIO challenging.
Even in high-speed driving conditions, sometimes exceeding 60 km/h, ARKit (red) shows surprisingly accurate and consistent 6-DoF motion tracking results overlaid on Google Maps as shown in the right of Fig.~\ref{figure_outdoor_campus_roads}.
The start and end points of ARKit (red) accurately meet in the black circle, and the final drift error (FDE) is only 2.68 m in Table~\ref{table_evaluation_FDE}.
ARCore (green) occasionally fails when the speed of the car increases or the light variations occur abruptly.
In T265 (blue), if the car stops temporarily due to a stop signal or is driving too fast, the VIO algorithm diverges and fails to estimate the location.
ZED2 (magenta) accumulates rotational drift error over time, resulting in inaccurate motion estimation results.
While four VIO systems perform relatively well in the previous walking sequences, this is not the case in the more challenging vehicular test, which is not officially supported by any of the tested VIO devices.
Only ARKit is able to produce stable motion tracking results even in a vehicular test.

We conduct an additional vehicular test driving the same trajectory repeatedly in a dark underground parking lot with poor visual conditions as shown in Fig.~\ref{figure_underground_parking_lot}.
The total traveling distance is about 450 m, and we drive the car at a low speed from 5 to 15 km/h.
Although ARKit does not restore the actual movements perfectly in the parking lot, ARKit (red) shows the overlapped, consistent motion estimation results while other VIO systems gradually diverge from the initially estimated loop.
Since we perform the evaluation at a relatively low speed (10 km/h) compared to the previous vehicle test (60 km/h), other VIO systems do not diverge or fail at all.
Among four VIO methods, the ZED2 positioning results are the most deviating from the actual movements in the underground parking lot.


\section{Discussion}
\label{sect:discussion}

Overall, {\bf Apple ARKit} demonstrates the most consistent, accurate, reliable, and stable motion tracking results among the four VIO systems across both indoor and outdoor uses.
ARKit performs well and robustly in various real-world challenging environments such as sudden camera movements, abrupt changes in illumination, and high-speed movements with very rare cases where tracking failure or motion jump occurs.
ARKit achieves accurate and robust positioning performance in realistic use cases where people and vehicles are very crowded not only indoors but also outdoors.

{\bf Google ARCore} exhibits accurate and consistent motion tracking performance next to ARKit.
ARCore works well for indoor sequences and the motion data collected by a walking person, but it diverges or the VIO algorithm deteriorates sharply when moving rapidly or in poor lighting conditions.

{\bf Intel RealSense T265} shows good positioning performance just behind Google ARCore.
T265 operates 6-DoF motion tracking indoors not badly, however, it has a problem of scale inconsistency issue when estimating the moving path larger or smaller than the scale of the actual movements.
Also, T265's motion tracking sometimes fails if the moving speed is too slow or fast.

The motion tracking performance of {\bf Stereolabs ZED 2} is the most inconsistent and inaccurate among the four VIO devices for indoors and outdoors.
As the 6-DoF motion tracking progresses, the rotational error occurs most severely, and this rotation error accumulates over time, resulting in an incorrect path in which the starting and ending points are very different.
In particular, ZED2 exhibits a tendency that it cannot track a straight path correctly when we actually move in a straight line outdoors, and rotational drift error is more severe when moving fast.

\section{Conclusion}
\label{sect:conclusion}

We have conducted a survey of the 6-DoF ego-motion tracking performance of four off-the-shelf proprietary VIO platforms in challenging indoor and outdoor environments.
To the best of our knowledge, this is the first back-to-back comparison of ARKit, ARCore, T265, and ZED2, demonstrating that Apple ARKit performs well and robustly in most indoor and outdoor scenarios.
We hope that the results and conclusions presented in this paper may help members of the research community in finding appropriate VIO platforms for their robotic systems and applications.


\bibliographystyle{IEEEtran}
\bibliography{pjinkim_ram}

\begin{thebibliography}{10}
\providecommand{\url}[1]{#1}
\csname url@rmstyle\endcsname
\providecommand{\newblock}{\relax}
\providecommand{\bibinfo}[2]{#2}
\providecommand\BIBentrySTDinterwordspacing{\spaceskip=0pt\relax}
\providecommand\BIBentryALTinterwordstretchfactor{4}
\providecommand\BIBentryALTinterwordspacing{\spaceskip=\fontdimen2\font plus
\BIBentryALTinterwordstretchfactor\fontdimen3\font minus
  \fontdimen4\font\relax}
\providecommand\BIBforeignlanguage[2]{{%
\expandafter\ifx\csname l@#1\endcsname\relax
\typeout{** WARNING: IEEEtran.bst: No hyphenation pattern has been}%
\typeout{** loaded for the language `#1'. Using the pattern for}%
\typeout{** the default language instead.}%
\else
\language=\csname l@#1\endcsname
\fi
#2}}

\bibitem{mourikis2007multi}
A.~I. Mourikis and S.~I. Roumeliotis, ``A multi-state constraint kalman filter
  for vision-aided inertial navigation,'' in \emph{IEEE ICRA}, 2007.

\bibitem{leutenegger2013keyframe}
S.~Leutenegger, P.~Furgale, V.~Rabaud, M.~Chli, K.~Konolige, and R.~Siegwart,
  ``Keyframe-based visual-inertial slam using nonlinear optimization,''
  \emph{Proceedings of Robotis Science and Systems (RSS) 2013}, 2013.

\bibitem{qin2018vins}
T.~Qin, P.~Li, and S.~Shen, ``Vins-mono: A robust and versatile monocular
  visual-inertial state estimator,'' \emph{IEEE Transactions on Robotics},
  2018.

\bibitem{apple2022arkit}
``{Apple} {ARKit},'' \url{https://developer.apple.com/documentation/arkit/},
  {A}ccessed: 2022-02-22.

\bibitem{google2022arcore}
``{Google} {ARCore},'' \url{https://developers.google.com/ar}, {A}ccessed:
  2022-02-22.

\bibitem{rouvcek2019darpa}
T.~Rou{\v{c}}ek, M.~Pecka, P.~{\v{C}}{\'\i}{\v{z}}ek,
  T.~Pet{\v{r}}{\'\i}{\v{c}}ek, J.~Bayer, V.~{\v{S}}alansk{\`y}, D.~He{\v{r}}t,
  M.~Petrl{\'\i}k, T.~B{\'a}{\v{c}}a, V.~Spurn{\`y}, \emph{et~al.}, ``Darpa
  subterranean challenge: Multi-robotic exploration of underground
  environments,'' in \emph{International Conference on Modelling and Simulation
  for Autonomous Systems}.\hskip 1em plus 0.5em minus 0.4em\relax Springer,
  2019.

\bibitem{root2021fast}
P.~Root, ``Fast lightweight autonomy (fla),'' \emph{Defense Advanced Research
  Projects Agency, https://www. darpa. mil/program/fast-lightweight-autonomy
  [retrieved 31 Dec. 2018]}, 2021.

\bibitem{flint2018visual}
A.~Flint, O.~Naroditsky, C.~P. Broaddus, A.~Grygorenko, S.~Roumeliotis, and
  O.~Bergig, ``Visual-based inertial navigation,'' Dec.~11 2018, {US Patent
  10,152,795}.

\bibitem{mourikis2009vision}
A.~I. Mourikis, N.~Trawny, S.~I. Roumeliotis, A.~E. Johnson, A.~Ansar, and
  L.~Matthies, ``Vision-aided inertial navigation for spacecraft entry,
  descent, and landing,'' \emph{IEEE Transactions on Robotics}, 2009.

\bibitem{nerurkar2020system}
E.~Nerurkar, S.~Lynen, and S.~Zhao, ``System and method for concurrent odometry
  and mapping,'' Oct.~13 2020, {US Patent 10,802,147}.

\bibitem{intel2022t265}
``{Intel RealSense Tracking Camera T265},''
  \url{https://www.intelrealsense.com/tracking-camera-t265/}, {A}ccessed:
  2022-02-22.

\bibitem{stereolabs2022zed2}
``{Stereolabs ZED 2 Stereo Camera},'' \url{https://www.stereolabs.com/zed-2/},
  {A}ccessed: 2022-02-22.

\bibitem{bloesch2015robust}
M.~Bloesch, S.~Omari, M.~Hutter, and R.~Siegwart, ``Robust visual inertial
  odometry using a direct ekf-based approach,'' in \emph{2015 IEEE/RSJ
  international conference on intelligent robots and systems (IROS)}, 2015.

\bibitem{mur2017orb}
R.~Mur-Artal and J.~D. Tard{\'o}s, ``Orb-slam2: An open-source slam system for
  monocular, stereo, and rgb-d cameras,'' \emph{IEEE transactions on robotics},
  2017.

\bibitem{engel2017direct}
J.~Engel, V.~Koltun, and D.~Cremers, ``Direct sparse odometry,'' \emph{IEEE
  transactions on pattern analysis and machine intelligence}, 2017.

\bibitem{delmerico2018benchmark}
J.~Delmerico and D.~Scaramuzza, ``A benchmark comparison of monocular {VIO}
  algorithms for flying robots,'' in \emph{IEEE ICRA}, 2018.

\bibitem{campos2021orb}
C.~Campos, R.~Elvira, J.~J.~G. Rodr{\'\i}guez, J.~M. Montiel, and J.~D.
  Tard{\'o}s, ``Orb-slam3: An accurate open-source library for visual,
  visual--inertial, and multimap slam,'' \emph{IEEE Transactions on Robotics},
  2021.

\bibitem{burri2016euroc}
M.~Burri, J.~Nikolic, P.~Gohl, T.~Schneider, J.~Rehder, S.~Omari, M.~W.
  Achtelik, and R.~Siegwart, ``The euroc micro aerial vehicle datasets,''
  \emph{The International Journal of Robotics Research}, vol.~35, no.~10, pp.
  1157--1163, 2016.

\bibitem{cortes2018advio}
S.~Cort{\'e}s, A.~Solin, E.~Rahtu, and J.~Kannala, ``{ADVIO}: An authentic
  dataset for visual-inertial odometry,'' in \emph{ECCV}, 2018.

\bibitem{alapetite2020comparison}
A.~Alapetite, Z.~Wang, and M.~Patalan, ``Comparison of three off-the-shelf
  visual odometry systems,'' \emph{Robotics}, 2020.

\bibitem{ouerghi2020comparative}
S.~Ouerghi, N.~Ragot, and X.~Savatier, ``Comparative study of a commercial
  tracking camera and {ORB-SLAM2} for person localization,'' in \emph{VISAPP},
  2020.

\bibitem{ling2018modeling}
Y.~Ling, L.~Bao, Z.~Jie, F.~Zhu, Z.~Li, S.~Tang, Y.~Liu, W.~Liu, and T.~Zhang,
  ``Modeling varying camera-imu time offset in optimization-based
  visual-inertial odometry,'' in \emph{Proceedings of the European Conference
  on Computer Vision (ECCV)}, 2018.

\bibitem{gumgumcu2019evaluation}
H.~Gümgümcü, ``Evaluation framework for proprietary slam systems exemplified
  on google arcore,'' Master's thesis, ETH Zurich, 2019.

\bibitem{nerurkar2014c}
E.~D. Nerurkar, K.~J. Wu, and S.~I. Roumeliotis, ``C-klam: Constrained
  keyframe-based localization and mapping,'' in \emph{2014 IEEE international
  conference on robotics and automation (ICRA)}.

\bibitem{marder2016project}
E.~Marder-Eppstein, ``Project tango,'' in \emph{ACM SIGGRAPH 2016 Real-Time
  Live!}, 2016, pp. 25--25.

\bibitem{bonatti2020learning}
R.~Bonatti, R.~Madaan, V.~Vineet, S.~Scherer, and A.~Kapoor, ``Learning
  visuomotor policies for aerial navigation using cross-modal
  representations,'' in \emph{2020 IEEE/RSJ International Conference on
  Intelligent Robots and Systems (IROS)}.

\bibitem{fan2019real}
R.~Fan, J.~Jiao, J.~Pan, H.~Huang, S.~Shen, and M.~Liu, ``Real-time dense
  stereo embedded in a uav for road inspection,'' in \emph{Proceedings of the
  IEEE/CVF Conference on Computer Vision and Pattern Recognition Workshops},
  2019.

\end{thebibliography}

\end{document}